\documentclass[conference]{IEEEtran}

\usepackage{cite}
\usepackage{amsmath,amssymb,amsfonts}
\usepackage{algorithmic}
\usepackage{graphicx}
\usepackage{textcomp}
\usepackage{xcolor}
\usepackage{hyperref}
\def\BibTeX{{\rm B\kern-.05em{\sc i\kern-.025em b}\kern-.08em
    T\kern-.1667em\lower.7ex\hbox{E}\kern-.125emX}}
\begin{document}

\title{FusionSORT: Fusion Methods for Online Multi-object Visual Tracking}

\author{{Nathanael L. Baisa} \\
\textit{School of Computer Science and Informatics} \\
\textit{De Montfort University}\\
Leicester LE1 9BH, UK \\
nathanael.baisa@dmu.ac.uk
}

\maketitle

\begin{abstract}
In this work, we investigate four different fusion methods for associating detections to tracklets in multi-object visual tracking. In addition to considering strong cues such as motion and appearance information, we also consider weak cues such as height intersection-over-union (height-IoU) and tracklet confidence information in the data association using different fusion methods. These fusion methods include minimum, weighted sum based on IoU, Kalman filter (KF) gating, and hadamard product of costs due to the different cues. We conduct extensive evaluations on validation sets of MOT17, MOT20 and DanceTrack datasets, and find out that the choice of a fusion method is key for data association in multi-object visual tracking. We hope that this investigative work helps the computer vision research community to use the right fusion method for data association in multi-object visual tracking. The source code is available at \textcolor{red}{\url{https://github.com/nathanlem1/FusionSORT}}.
\end{abstract}


\section{Introduction}

Multi-object visual tracking is currently an active research field in computer vision due to its wide range of applications including, but not limited to, intelligent surveillance, autonomous driving, robot navigation and augmented reality. Its main goal is to detect objects and recognize their identities in video stream in order to produce their trajectories. The most commonly adopted paradigm for multi-object visual tracking in computer vision is tracking-by-detection~\cite{BewGeOtt16}\cite{WojBewPau17}\cite{DuZhiSon23}\cite{zhaSunJia22}\cite{NirRoyBen22}\cite{SeiBarCas23}\cite{BAISA21}. This is a result of the significant progress achieved in object detection algorithms powered by deep learning. In this tracking-by-detection paradigm, object detections are first obtained from object detector applied to video frames, which is considered as a detection step. This is then followed by a tracking step, where state estimation and data association are conducted. The motion prediction of a state estimation~\cite{WelBis06} predicts the bounding boxes of object tracklets in the next frame, and then the data association is performed between these tracklets and current detections to update the tracklets for generating trajectories of tracked objects over-time. The standard choice of the state estimation method for multi-object visual tracking is Kalman filter (KF)~\cite{WelBis06}.

Data association is very challenging in multi-object visual tracking due to challenges such as miss-detections due to occlusions, appearance changes, or noisy detections. To solve the association task between the predicted tracklet bounding box and the detection bounding box, many works use either one or a combination of strong cues, motion and appearance information, since these cues provide powerful instance-level discrimination. The motion information is usually computed using intersection-over-union (IoU) and its variants~\cite{Stadler2023} or Mahalanobis distance~\cite{MaeJouMas20}. Normalized Euclidean distance between bounding box centers of detection and tracklet is also used in~\cite{BAISA21}. The appearance information is usually leveraged from trained deep learning models~\cite{LinXinWu23}\cite{BAISA2024}. The faster trackers such as~\cite{BewGeOtt16}\cite{zhaSunJia22}\cite{CaoPanWen23} use only the motion information for the association task. Many trackers use a combination of motion and appearance information~\cite{WojBewPau17}\cite{DuZhiSon23}\cite{NirRoyBen22}\cite{MagAhmCao23}\cite{WanZheLiu20}\cite{ZhaWanWan21}\cite{DanJur23}, generally with better performance but with compromised speed. In addition to the strong cues, some works also incorporate weak cues such as height state, confidence state and/or velocity direction for the association task~\cite{YanHanYan24}\cite{YuLiLi16}\cite{CaoPanWen23} to compensate for the strong cues, especially in challenging situations such as occluded and crowded scenes. The key gap that is missing in the literature is the comparative study on the effectiveness of the strategies for fusing different cues such as strong cues and/or weak cues. Different trackers use different fusion methods; however, there is no work in the literature, to the best of our knowledge, which investigates and compares the different fusion methods for data association in multi-object visual tracking. 

In this work, we investigate different fusion methods used in multi-object visual tracking and evaluate them extensively on different tracking datasets. Our tracker obeys the Simple, Online and Real-Time (SORT) characteristics; hence, we call our tracker FusionSORT. We design our tracker in such a way that we can flexibly use the different strong cues and/or weak cues with different fusion methods for the thorough investigation of our tracker through extensive experiments. Moreover, we elegantly incorporate tracklet confidence state into the state vector representation of the KF. In general, the main contributions of this paper are as follows:

\begin{enumerate}
  \item We investigate four widely used different fusion methods for associating detections to tracklets in multi-object visual tracking, including minimum, weighted sum based on IoU, KF gating, and hadamard product of costs.
  \item In addition to considering strong cues such as motion and appearance information, we also incorporate weak cues such as height-IoU and tracklet confidence information in the data association.
  \item We conduct extensive evaluations on validation sets of MOT17, MOT20 and DanceTrack datasets, and demonstrate that the choice of a fusion method is key for data association in multi-object visual tracking.
\end{enumerate}

The rest of this paper is organized as follows. After the discussion of related work in section~\ref{RelatedWork}, our proposed method is explained in detail including strong and weak cues modelling and fusion methods in section~\ref{Method}. The experimental setting and results are analyzed and compared in section~\ref{Experiments}, followed by the main conclusion in section~\ref{Conclusion}.

\section{Related Work}\label{RelatedWork}


We give an overview of related work on tracking-by-detection and data association.

\subsection{Tracking-By-Detection}

Tracking-by-detection is the most widely adopted paradigm for multi-object visual tracking in computer vision~\cite{BewGeOtt16}\cite{WojBewPau17}\cite{DuZhiSon23}\cite{zhaSunJia22}\cite{NirRoyBen22}\cite{SeiBarCas23}\cite{BAISA21}. This is due to the significant progress achieved in object detection algorithms driven by deep learning. There are two steps in this tracking-by-detection paradigm: detection and tracking. In the detection step, object detection bounding boxes are obtained by applying object detector to video frames. This is then followed by a tracking step, where state estimation and data association are accomplished. Kalman filter (KF)~\cite{WelBis06} with a constant-velocity model for motion estimation is the commonly used state estimation method for multi-object visual tracking~\cite{BewGeOtt16}\cite{WojBewPau17}\cite{CheAiZhu18} due to its simplicity and efficiency. A Gaussian mixture probability hypothesis density (GM-PHD) filter has also been used in many works for multi-object visual tracking~\cite{BAISA21}\cite{Baisa19}\cite{ZeyFedSye18}\cite{BaiWal19}. These trackers have separate detection and tracking components. Recently, several joint trackers~\cite{WanZheLiu20}\cite{ZhaWanWan21}\cite{RenHanDin23}\cite{YouYaoBao23} have been proposed which jointly train detection and some other components such as motion, embedding and association models. The primary advantage of these joint trackers is their low computational requirements combined with similar performance levels.

\subsection{Data Association}
Data association in multi-object visual tracking is highly challenging due to factors like missed detections caused by occlusions, changes in object appearance, and noisy input detections. Identifying the temporally stable properties of objects is crucial to effectively associate predicted tracklet bounding boxes to detection bounding boxes in video frames. Strong cues such as motion and appearance information provide powerful instance-level discrimination. The motion information is usually computed using IoU and its variants~\cite{Stadler2023} or Mahalanobis distance~\cite{MaeJouMas20}. Normalized Euclidean distance between bounding box centers of detection and tracklet is also used in~\cite{BAISA21}. The appearance information is usually leveraged from trained deep learning models~\cite{LinXinWu23}\cite{BAISA2024}. Specifically, deep appearance features are extracted from image patches determined by object detection boxes using an additional deep neural network in separate appearance-based trackers~\cite{WojBewPau17}\cite{DuZhiSon23}\cite{zhaSunJia22}\cite{NirRoyBen22}. Appearance models can also be trained jointly with object detectors in joint trackers~\cite{WanZheLiu20}\cite{ZhaWanWan21}\cite{RenHanDin23}\cite{YouYaoBao23}. For the association task, cosine distance of the extracted deep appearance features is computed as appearance distance. Weak cues such as height state, confidence state and/or velocity direction can provide informative clues that help to compensate for the discrimination of strong cues such as motion and appearance information for associating predicted tracklet boxes to new detection boxes~\cite{YanHanYan24}\cite{YuLiLi16}\cite{CaoPanWen23}. 

Combining these different sources of information for associating predicted tracket boxes to new detection boxes is very crucial. Though several trackers use only motion information~\cite{BewGeOtt16}\cite{zhaSunJia22}\cite{CaoPanWen23} for data association with interesting performance, the performance can be improved by fusing different cues. Minimum fusion method is used in~\cite{NirRoyBen22}, where the minimum in each element of the cost matrices of motion and appearance is used to match tracklets to new detections. Several works~\cite{DanJur23}\cite{YanHanYan24}\cite{HasKoiHwa24} use weighted sum, with some variations, of motion and appearance information, in which motion information is computed using IoU. Another fusion method, which we call KF gating, is also based on weighted sum of cues and is used in~\cite{WanZheLiu20}\cite{ZhaWanWan21}\cite{DuZhiSon23}. However, the Mahalanobis distance is used in this fusion method instead of the IoU, and is subjected to KF gating. Hadamard product is also used in~\cite{ZeyFedSye18}\cite{YuLiLi16} which computes element-wise multiplication of different costs. Hence, different trackers use different fusion methods to fuse different costs for associating tracklets to current detections, which is then solved by Hungarian algorithm~\cite{Kuhn1955} as bipartite graph matching. However, there is no work in the literature which thoroughly investigates these different fusion methods. In this work, we investigate four commonly used fusion methods and demonstrate that the choice of a fusion method is key for data association in multi-object visual tracking.

\section{Method}\label{Method}
In our tracker, we use Kalman filter (KF)~\cite{WelBis06} with a constant-velocity model for motion estimation of object tracklets in the image plane, similar to the other SORT methods~\cite{BewGeOtt16}\cite{WojBewPau17}\cite{zhaSunJia22}\cite{NirRoyBen22}. The cost matrices are computed by measuring the pairwise representation similarity between tracklets and detections for the association task, which is then solved by Hungarian algorithm~\cite{Kuhn1955} as bipartite graph matching. For computing the total cost matrix, we consider strong cues such as motion and appearance information as well as weak cues such as height-IoU and confidence information. Furthermore, our tracker incorporates camera-motion compensation (CMC), as used in~\cite{NirRoyBen22}\cite{PhiTimLau19}. We adopted the two stage matching strategy, similar to previous works~\cite{zhaSunJia22}\cite{NirRoyBen22}, that conducts the first association using high-confident detections and then the second association using low-confident detections. Appearance information, and hence, the fusion methods are used only at the first association stage. The second association stage matches the low-confident detections to the remaining unassigned tracklets that have not been assigned to high-confident detections in the first association stage. In addition, the second association uses only the motion information, specifically IoU. Only the high-confident detections with scores above a given threshold are used for new track initialization. The low-confident detections are not utilized to start new tracks in order to avoid false-positive tracks that can be introduced from low-confident false positive detections.

For state vector representation, we extend the widely used standard KF in~\cite{NirRoyBen22} with two additional states: the tracklet confidence (score) $c$ and its velocity component $\dot{c}$, following the state vector derivation approach in~\cite{baisa20}. Accordigly, the state vector is represented as in~\eqref{state}

\begin{equation}
\begin{array} {lll}
  \mathbf{x}_k =& [x_c(k), y_c(k), w(k), h(k), c(k), \\ &
                 \dot{x}_c(k), \dot{y}_c(k), \dot{w}(k), \dot{h}(k), \dot{c}(k)]^T
\end{array}
\label{state}
\end{equation}
\noindent where $(x_c, y_c)$ denote object tracklet box’s center, while $w$, $h$ and $c$ represent the object tracklet box’s width, height, and tracklet confidence, respectively. The velocity components are denoted by $\dot{x_c}$, $\dot{y_c}$, $\dot{w}$, $\dot{h}$ and $\dot{c}$. Tracklet box size can change dramatically when predicting inactive (lost) tracks for multiple frames without state update, which hinders re-activation after occlusion. To overcome this, we apply height preservation, setting the derivative $\dot{h}$ to zero before the KF prediction step, similar to~\cite{Stadler2023}\cite{NirRoyBen22}\cite{zhaSunJia22}. Similarly, we also apply width preservation ($\dot{w} = 0$) and tracklet confidence preservation ($\dot{c} = 0$) before the KF prediction step in our experiments. Note that the tracklet confidence preservation has less impact on some sequences when compared to the others.

Similarly, the measurement vector is represented as in~\eqref{measurement}

\begin{equation}
  \mathbf{z}_k = [z_{xc}(k), z_{yc}(k), z_w(k), z_h(k), z_c(k)]^T
\label{measurement}
\end{equation}
\noindent where $(z_{xc}, z_{yc})$ denote object detection box’s center, while $z_w$, $z_h$ and $z_c$ represent the object detection box’s width, height and score, respectively.

Following the extension of the above state and measurement vectors, we also extend the process noise covariance $\mathbf{Q}_k$ and the measurement noise covariance $\mathbf{R}_k$ matrices as in~\eqref{Qk} and~\eqref{Rk}, respectively, which incorporate the tracklet confidence $c$ and its velocity component $\dot{c}$. Following~\cite{WojBewPau17}\cite{NirRoyBen22}, we use time-dependent $\mathbf{Q}_k$ and $\mathbf{R}_k$ which are expressed as functions of some estimated elements and some measurement elements.

\begin{equation}
\begin{array} {lll}
  \mathbf{Q}_k =& diag((\sigma_p \hat{w}_{k-1|k-1})^2, (\sigma_p \hat{h}_{k-1|k-1})^2, \\& 
                 (\sigma_p \hat{w}_{k-1|k-1})^2, (\sigma_p \hat{h}_{k-1|k-1})^2, \\& 
                 (\sigma_p \hat{c}_{k-1|k-1})^2, (\sigma_v \hat{w}_{k-1|k-1})^2, \\& 
                 (\sigma_v \hat{h}_{k-1|k-1})^2, (\sigma_v \hat{w}_{k-1|k-1})^2, \\& 
                 (\sigma_v \hat{h}_{k-1|k-1})^2, (\sigma_v \hat{c}_{k-1|k-1})^2)
\end{array}
\label{Qk}
\end{equation}
\noindent

\begin{equation}
\begin{array} {lll}
  \mathbf{R}_k =& diag((\sigma_m \hat{w}_{k|k-1})^2, (\sigma_m \hat{h}_{k|k-1})^2, \\& 
                 (\sigma_m \hat{w}_{k|k-1})^2, (\sigma_m \hat{h}_{k|k-1})^2, (\sigma_m \hat{c}_{k|k-1})^2)
\end{array}
\label{Rk}
\end{equation}
\noindent

Following the works in~\cite{NirRoyBen22}\cite{WojBewPau17}, we choose the noise factors as $\sigma_p = 0.05$, $\sigma_v = 0.00625$, and $\sigma_m = 0.05$, since our frame rate is also 30 fps. It is worth noting that we modified $\mathbf{Q}_k$ and $\mathbf{R}_k$ according to our slightly modified state vector $\mathbf{x}_k$ and measurement vector $\mathbf{z}_k$, respectively. Noise Scale Adaptive (NSA) KF~\cite{DuZhiSon23}, $\mathbf{R}_{NSA} = (1 - z_c)\mathbf{R}$, did not help in our experiments. Ideally, the higher the detection confidence, the smaller the adapted measurement noise
covariance $\mathbf{R}_{NSA}$ and the more influence has the detection on the track state update.

\subsection{Strong Cues}

The association task in multi-object visual tracking is primarily solved, explicitly or implicitly, by using strong cues such as motion and appearance information since these cues provide powerful instance-level discrimination. In this work, we consider both intersection-over-union (IoU) and Mahalanobis distance~\cite{MaeJouMas20} as motion information.

Given two boxes as $b^1 = (x^1_1, y^1_1, x^1_2, y^1_2)$ and $b^2 = (x^2_1, y^2_1, x^2_2, y^2_2)$, where $x_1$ and $y_1$ represents the top-left corner and  $x_2$ and $y_2$  represents the bottom-right corner, the conventional IoU based on area can be given as in~\eqref{iou}.

\begin{equation}
  IoU = \frac{B_1 \cap B_2}{B_1 \cup B_2}
\label{iou}
\end{equation}
\noindent where $B_1$ and $B_2$ are the areas of the boxes $b^1$ and $b^2$, respectively.

It is possible to incorporate the uncertainty of the motion estimation into the distance measure since the KF is used as a motion model. Given a probability distribution $f$ on $\mathcal R^N$, with mean $\mathbf{\mu} = (\mu_1, \mu_2, \mu_3, \dots, \mu_N)^\mathsf{T}$ and positive semi-definite covariance matrix $S$, the Mahalanobis distance of a point $\mathbf{z} = (z_1, z_2, z_3, \dots, z_N )^\mathsf{T}$ from $f$ is generally given as in~\eqref{mahalanobis}

\begin{equation}
  d_M(\mathbf{z}, f) = \sqrt{(\mathbf{z} - \mathbf{\mu})^\mathsf{T} S^{-1} (\mathbf{z} - \mathbf{\mu})}
\label{mahalanobis}
\end{equation}
\noindent where $\mathbf{z}$ and $\mathbf{\mu}$ correspond to measurement (detection) box center position and the projection of the estimated tracklet mean into measurement space, respectively, while $f$ corresponds to a Gaussian predicted state distribution. $S^{-1}$ denotes inverse of the projected tracklet state covariance matrix $S$ into the measurement space. Hence, ($\mathbf{\mu}$, $S$) corresponds to a track state projected into the measurement space. We use the squared Mahalanobis distance $d^2_M(\mathbf{z}, f)$ in our experiments.

For appearance information, we exploited deep appearance representation, particularly using stronger baseline on top of BoT (SBS)~\cite{HaoYouXin19} with the ResNeSt50~\cite{HanChoZho22} as a backbone, from the FastReID library~\cite{LinXinWu23}, as used in~\cite{NirRoyBen22}\cite{MagAhmCao23}. For updating the matched tracklet appearance embedding $e^k_i$ for the i-th tracklet at frame k, we use the exponential moving average (EMA) method, similar to~\cite{WanZheLiu20}\cite{NirRoyBen22}, as given in~\eqref{ema}.

\begin{equation}
  e^k_i = \alpha e^{k-1}_i  +  (1- \alpha) f^k_i
\label{ema}
\end{equation}
\noindent where $\alpha = 0.9$ is a momentum term and $f^k_i$ is the appearance embedding of the current matched detection. The appearance features are extracted only from the high-confident detections i.e. appearance features are used only in the first association step. We compute cosine similarity between the averaged tracklet appearance embedding vector $e^k_i$ and the new detection embedding vector $f^k_i$ to match them.

\subsection{Weak Cues}

Though strong cues are the widely used information for associating detections to tracklets in multi-object visual tracking, they suffer from degradation under challenging situations such as occluded and crowded scenes~\cite{YanHanYan24}. Hence, we employ weak cues such as confidence and height state to compensate for the strong cues. For incorporating height information, we compute height intersection-over-union (hIoU) given tracklet and detection bounding boxes.

Accordingly, given the two boxes as $b^1$ and $b^2$ above, the height-IoU (hIoU) can be computed as in~\eqref{hiou}

\begin{equation}
  hIoU = \frac{min(y^1_2, y^2_2) - max(y^1_1, y^2_1)}{max(y^1_2, y^2_2) - min(y^1_1, y^2_1)}
\label{hiou}
\end{equation}
\noindent

We compute the confidence cost as the absolute difference between the estimated tracklet confidence $c_{trk}$ and detection confidence $c_{det}$, as given in~\eqref{confidence}.

\begin{equation}
  C_c = |c_{trk} - c_{det}|
\label{confidence}
\end{equation}
\noindent where the tracklet confidence $c_{trk}$ is given in~\eqref{state} as $c(k)$, and $c_{det}$ is the detection score obtained from object detector applied to a video frame. Refer to~\eqref{confidence-disance} for explicit formulation of confidence cost.

\subsection{Fusion Methods} \label{FusionMethods}
The four fusion methods that we use in our investigative experimental analysis are described as follows. Note that the fusion methods are applied only for the first association step; the second association step uses only IoU for all the experiments.

\subsubsection{Minimum}

In this fusion method, we use the \textit{minimum} in each element of the cost matrices of motion, appearance, height-IoU and confidence as the final value of the cost matrix $C$. IoU is used for computing the motion cost. We extend the minimum fusion method designed for motion and appearance costs in~\cite{NirRoyBen22} to include the height-IoU and confidence costs as in~\eqref{minimum-cos},~\eqref{minimum-hiou},~\eqref{minimum-conf} and~\eqref{minimum}.

\begin{equation}
    \hat{d}^{cos}_{i,j} =
\begin{cases}
    0.5~.~d^{cos}_{i,j} (d^{cos}_{i,j} < \theta_{emb}) \wedge (d^{iou}_{i,j} < \theta_{iou})  \\
    1,        ~\text{otherwise}
\end{cases}
\label{minimum-cos}
\end{equation}
\noindent

\begin{equation}
    \hat{d}^{hiou}_{i,j} =
\begin{cases}
    d^{hiou}_{i,j} \wedge (d^{iou}_{i,j} < \theta_{iou})  \\
    1,        ~\text{otherwise}
\end{cases}
\label{minimum-hiou}
\end{equation}
\noindent

\begin{equation}
    \hat{d}^{conf}_{i,j} =
\begin{cases}
    d^{conf}_{i,j} \wedge (d^{iou}_{i,j} < \theta_{iou})  \\
    1,        ~\text{otherwise}
\end{cases}
\label{minimum-conf}
\end{equation}
\noindent

\begin{equation}
  C_{i,j} = min(\hat{d}^{cos}_{i,j}, d^{iou}_{i,j}, \hat{d}^{hiou}_{i,j}, \hat{d}^{conf}_{i,j})  
\label{minimum}
\end{equation}
\noindent where $C_{i,j}$ denotes the (i, j) element of the total cost matrix $C$. $d^{cos}_{i,j}$, $d^{iou}_{i,j}$, $d^{hiou}_{i,j}$ and $d^{conf}_{i,j}$ are the cosine distance, IoU distance, height-IoU distance and confidence distance between i-th tracklet and j-th detection, respectively. $\wedge$ denotes logical 'and'. The minimum fusion is represented by \textit{min} in~\eqref{minimum}. Note that cosine distance is 1 minus cosine similarity, and the other distances generally follow similar manner, and they are explicitly formulated as in~\eqref{cosine-disance},~\eqref{iou-disance},~\eqref{hiou-disance}, and~\eqref{confidence-disance}.

\begin{equation}
  d^{cos}_{i,j} = 1 - \frac{e^k_i . f^k_j}{\Vert e^k_i\Vert_2 \Vert f^k_j \Vert_2}
\label{cosine-disance}
\end{equation}
\noindent where $\Vert * \Vert_2$ is the 2-norm of its argument *, and . represents the dot product of tracklet averaged appearance embedding $e^k_i$ and detection appearance embedding $f^k_j$. Note that $d^{cos}_{i,j} \in [0,2]$.

\begin{equation}
  d^{iou}_{i,j} = 1 - IoU_{i,j}
\label{iou-disance}
\end{equation}
\noindent $d^{iou}_{i,j} \in [0,1]$ is IoU distance, also called Jaccard distance.

\begin{equation}
  d^{hiou}_{i,j} = 1 - hIoU_{i,j}
\label{hiou-disance}
\end{equation}
\noindent

\begin{equation}
  d^{conf}_{i,j} = |c^{trk}_i - c^{det}_j|
\label{confidence-disance}
\end{equation}
\noindent where $|*|$ is the absolute value of *. $c^{trk}_i$ and $c^{det}_j$ denote tracklet confidence and detection score, respectively. Note that $d^{hiou}_{i,j} \in [0,1]$ and $d^{conf}_{i,j} \in [0,1]$.

Accordingly, cost matrices $C_a$, $C_m$, $C_h$ and $C_c$ are constructed from $d^{cos}_{i,j}$, $d^{iou}_{i,j}$, $d^{hiou}_{i,j}$ and $d^{conf}_{i,j}$ to represent the appearance, motion, height-IoU and confidence costs, respectively. The IoU threshold $\theta_{iou}$ and appearance threshold $\theta_{emb}$ are set to 0.5 and 0.25, respectively. These thresholds are used to discard low cosine similarity or far away candidates in terms of IoU’s value.

\subsubsection{Weighted Sum}
Given the appearance cost $C_a$, motion cost $C_m$, height-IoU cost $C_h$ and confidence cost $C_c$, this method uses weighted sum for calculating the total cost matrix $C$ as in \eqref{weightedsum}. In this method, the motion cost is computed using IoU. 

\begin{equation}
  C = \lambda_1 C_m + \lambda_2 C_a + \lambda_3 C_h + \lambda_4 C_c  
\label{weightedsum}
\end{equation}
\noindent where the $\lambda_1$,  $\lambda_2$, $\lambda_3$ and  $\lambda_4$ are the weights for motion, appearance, height-IoU and confidence costs, respectively. We used the appearance threshold similar to~\eqref{minimum-cos}; however, we did not apply the IoU threshold mask for the $C_h$ and $C_c$ costs as in~\eqref{minimum-hiou}~\eqref{minimum-conf} since it slightly decreases the results. We set $\lambda_1$, $\lambda_3$ and  $\lambda_4$ to 1.0, 0.1 and 0.1, respectively, for all of the datasets.   $\lambda_2$ is set to 0.1 for MOT17 and MOT20 datasets and 0.2 for DanceTrack dataset.

\subsubsection{KF Gating}

Given the appearance cost $C_a$, motion cost $C_m$, height-IoU cost $C_h$ and confidence cost $C_c$, this method also uses weighted sum for calculating the total cost matrix $C$ as in \eqref{KFgating}. However, the motion cost is computed using Mahalanobis distance, rather than IoU,  and it is subjected to KF gating~\cite{WelBis06} (preventing unlikely assignments) where the gating distance is computed between KF predicted state distribution and measurements (detections). A suitable Mahalanobis distance threshold can be obtained from a table for the 0.95 quantile of the chi-square distribution with N degrees of freedom. In our method, the chi-square distribution has 2 degrees of freedom since the distance computation is done with respect to the bounding box center position only i.e. ($x_c, y_c$).

\begin{equation}
  C = \lambda (C_a + \lambda_h C_h + \lambda_c C_c)  +  (1- \lambda) C_m
\label{KFgating}
\end{equation}
\noindent where the weight factor $\lambda$ is set to 0.98, and the $\lambda_h$ and $\lambda_c$ are the weights for the height-IoU and confidence costs, respectively. Both $\lambda_h$ and $\lambda_c$ are set to 0.2.

Even though the Mahalanobis distance is used for the first association step where fusion of different costs takes place, the IoU is used for the second association step for all the fusion methods including this fusion method to make a fair comparison. In our experiments, the IoU gives better overall performance than the Mahalanobis distance for the second association step, as shown in Table~\ref{IoUvsMahalanobis}. Note that we did not apply any thresholding and 0.5 multiplication as in~\eqref{minimum-cos},~\eqref{minimum-hiou}, and~\eqref{minimum-conf} in this fusion method.

\subsubsection{Hadamard Product}
In this fusion method, the total cost matrix $C$ is obtained by element-wise multiplication of all costs: appearance cost $C_a$, motion cost $C_m$, height-IoU cost $C_h$ and confidence cost $C_c$, as in \eqref{hadamard}. In this method, the motion cost is computed using IoU. 

\begin{equation}
  C = C_a \odot C_m \odot C_h \odot C_c
\label{hadamard}
\end{equation}
\noindent
We used the appearance and IoU thresholds in a similar manner as in~\eqref{minimum-cos},~\eqref{minimum-hiou} and~\eqref{minimum-conf}, which improve the results.

\section{Experiments} \label{Experiments}

\subsection{Experimental Setting}

\subsubsection{Datasets}
We conduct our investigative experiments on different multi-object visual tracking benchmarks, including MOT17~\cite{MilLeaRei16}, MOT20~\cite{DenRezMil20} and DanceTrack~\cite{SunCaoJia22}, which were captured under diverse scenarios. MOT17 was captured using both static and moving cameras and consists of seven train and seven test sequences, in which the motion is mostly linear. MOT20 consists of highly crowded four train and four test sequences which are used to evaluate trackers under dense objects and severe occlusions. DanceTrack stands out as one of the most demanding tracking benchmarks, characterized by a variety of non-linear motion patterns, frequent interactions, and significant occlusions. It contains 40, 25 and 35 videos of dancing humans for training, validation and testing. The MOT17 and MOT20 validation sets follow a widely adopted convention~\cite{ZhoKolKra20}\cite{zhaSunJia22}\cite{NirRoyBen22}\cite{MagAhmCao23}\cite{YanHanYan24} where the train set is split into halves for training and validation since these datasets do not have a separate validation set i.e. they have only train and test sets. The DanceTrack has a separate validation set, in addition to train and test sets, which we use directly. Hence, our experimental analysis is based on the validation sets of these benchmarks.

\subsubsection{Evaluation Metrics}
We use different evaluation metrics for comparing tracking performance based on the different fusion methods, including 
Multiple Object Tracking Accuracy (MOTA)~\cite{BerSti08}, Identification F1 (IDF1)~\cite{RisSolZou16} and Higher-Order Tracking Accuracy (HOTA)~\cite{JonAljPat21}. MOTA mainly focuses on evaluating the detection performance while IDF1 evaluates the identity association performance of a tracker. HOTA combines several sub-metrics that evaluate the tracker from different perspectives, providing a comprehensive assessment of the tracker performance, including detection, association, and localization into a single unified metric. 

\subsubsection{Implementation Details}
Our visual tracking algorithm is implemented using Python and PyTorch deep learning framework, and run on Laptop with Intel(R) Core(TM) i7-10850H @ 2.70GHz, 16 GB RAM and NVIDIA GeForce RTX 2070 GPU. We use the publicly available YOLOX-X detector~\cite{GeLiuWan21}, trained by~\cite{zhaSunJia22} for MOT17 and MOT20 datasets, and trained by~\cite{YanHanYan24} for DanceTrack dataset. For feature extraction, we used the publicly available models trained by~\cite{NirRoyBen22} for MOT17 and MOT20 datasets and trained by~\cite{YanHanYan24} for DanceTrack dataset, based on FastReID library~\cite{LinXinWu23}. We use the same tracker parameters throughout our experimental analysis, which were mostly set empirically. Unless otherwise specified, we set high detection score threshold $\tau_1$ to 0.6, which is used to separate high-confident detections from low-confident detections for the first association step. We set low detection score threshold $\tau_2$ to 0.1 for the second association step. Detections with score lower than $\tau_2$ are discarded. In the linear assignment step, the matching is rejected if the detection and the tracklet similarity is smaller than 0.2 for the first association step and smaller than 0.5 for the second association step. The detection score needs to be at least 0.7 to be considered for track initialization. The lost tracks (inactive tracks) are kept for 30 frames in case they appear again before they get deleted. Note that we do not output the boxes and identities of lost tracks, as in~\cite{zhaSunJia22}\cite{NirRoyBen22}. In order to precisely investigate the effectiveness of the fusion methods, we do not apply any tracklet interpolation as a post-processing in our experiments.

\subsection{Experimental Results}

We compare FusionSORT on the validation set of MOT17, MOT20 and DanceTrack using different fusion methods in Table~\ref{MOT17}, Table~\ref{MOT20} and Table~\ref{DanceTrack}, respectively.

\subsubsection{MOT17}

Different fusion methods are compared on validation set of MOT17 in Table~\ref{MOT17}. As shown in this table, the highest values of HOTA, MOTA and IDF1 are obtained using minimum, weighted sum based on IoU and KF gating, respectively. Fusing appearance information to motion information generally improves performance when using minimum, weighted sum based on IoU and KF gating methods. However, the performance degrades when using hadamard fusion method since the hadamard fusion method treats both motion and appearance information with equal importance implicitly, where the contribution of the motion information is higher in this case. The incorporation of weak cues such as height-IoU and confidence information negatively affects the tracking performance when using the minimum fusion method. This happens because the minimum fusion method does not fully exploit the potential of all the cues since one of them is used for the association task, where the weak cues are expected to contribute minimally when compared to the strong cues. Similarly, significant performance decline is observed when fusing the weak cues with the strong cues using the hadamard fusion method since it implicitly treats all cues equally. The incorporation of weak cues improves the tracking performance when using the KF gating fusion method. Note that all the fusion methods use IoU for computing motion distance except the KF gating method which uses Mahalanobis distance.

\begin{table}[htbp]
\caption{Evaluation on MOT17 validation dataset. Motion distance (mot) is computed using Mahalanobis distance for KF gating fusion method and using IoU for the other methods. The first and second highest values are highlighted by $\color{red}{\text{red}}$ and $\color{blue}{\text{blue}}$, respectively.}
\begin{center}
\begin{tabular}{|c|c|c|c|}
\hline
\textbf{ }&\multicolumn{3}{|c|}{\textbf{MOT17}} \\
\cline{2-4} 
\textbf{Minimum} &  \textbf{\textit{MOTA}}& \textbf{\textit{IDF1}}& \textbf{\textit{HOTA}} \\
\hline 
mot (iou) & 78.421 & 81.999 & 69.345 \\
\hline 
mot, app  & \color{blue}78.541 & 82.238 & \color{red}69.419 \\
\hline 
mot, app, hiou  & 78.209 & 80.937 & 68.586 \\
\hline 
mot, app, hiou, confidence  & 78.079 & 79.556 &	66.896 \\ 
\hline
\textbf{Weighted-sum}&\multicolumn{3}{|c|}{ } \\
\cline{2-4} 
\hline
mot (iou) & 78.421 & 81.999 & 69.345  \\
\hline 
mot, app & \color{red}78.584 &	81.877	& \color{blue}69.379 \\
\hline 
mot, app, hiou  & \color{blue}78.541 & 81.524 & 69.246  \\
\hline 
mot, app, hiou, confidence  &  78.432 &	81.15 &	68.812  \\
\hline
\textbf{KF-gating}&\multicolumn{3}{|c|}{ } \\
\cline{2-4} 
\hline
mot (mahalanobis)  &  76.749 & 69.256 &  60.581 \\
\hline 
mot, app  & 78.035 &  81.75 &  68.889 \\
\hline 
mot, app, hiou  & 78.072  &  \color{blue}82.249 &  69.29 \\
\hline 
mot, app, hiou, confidence  &  78.003 & \color{red}82.322 &  69.366 \\
\hline
\textbf{Hadamard}&\multicolumn{3}{|c|}{ } \\
\hline
mot (iou) & 78.421 & 81.999 & 69.345  \\
\hline 
mot, app  &  78.311 &  80.718 & 68.38 \\
\hline 
mot, app, hiou  &  78.282 & 80.415 & 68.172 \\
\hline 
mot, app, hiou, confidence  & 78.261  & 80.045 & 67.882 \\
\hline
\end{tabular}
\label{MOT17}
\end{center}
\end{table}

\subsubsection{MOT20}
As can be seen in Table~\ref{MOT20}, all fusion methods improve the tracking performance when we combine the appearance information with the motion information. Incorporating the weak cues such as height-IoU and confidence information improve the performance when using the weighted sum based on IoU and KF gating fusion methods. The highest values of HOTA, MOTA and IDF1 are obtained with the KF gating fusion method by combining both strong cues such as motion and appearance information and weak cues such as height-IoU and confidence information. However, combining weak cues with the strong cues degrades the tracking performance when using minimum and hadamard fusion methods.

\begin{table}[htbp]
\caption{Evaluation on MOT20 validation dataset. Motion distance (mot) is computed using Mahalanobis distance for KF gating fusion method and using IoU for the other methods. The first and second highest values are highlighted by $\color{red}{\text{red}}$ and $\color{blue}{\text{blue}}$, respectively.}
\begin{center}
\begin{tabular}{|c|c|c|c|}
\hline
\textbf{ }&\multicolumn{3}{|c|}{\textbf{MOT20}} \\
\cline{2-4} 
\textbf{Minimum} &  \textbf{\textit{MOTA}}& \textbf{\textit{IDF1}}& \textbf{\textit{HOTA}} \\
\hline 
mot (iou) & 72.722 & 73.794 & 57.815 \\
\hline 
mot, app  & 72.714	& 74.232 &	58.216 \\
\hline 
mot, app, hiou  & 72.701 &	73.653 & 57.69 \\
\hline 
mot, app, hiou, confidence  & 72.442	& 70.809 & 55.764  \\
\hline
\textbf{Weighted-sum}&\multicolumn{3}{|c|}{ } \\
\cline{2-4} 
\hline
mot (iou) &  72.722 & 73.794	& 57.815 \\
\hline 
mot, app  & 72.682	& 74.505 &	58.262 \\
\hline 
mot, app, hiou  & 72.674 &	74.536 & 58.275  \\
\hline 
mot, app, hiou, confidence & 72.706	& 74.626 & 58.368 \\
\hline
\textbf{KF-gating}&\multicolumn{3}{|c|}{ } \\
\cline{2-4} 
\hline
mot (mahalanobis) & 71.932 & 60.222 & 48.78 \\
\hline 
mot, app  & \color{red}73.052	& 74.502 & 58.236  \\
\hline 
mot, app, hiou  & \color{blue}72.993 & \color{blue}74.724 & \color{blue}58.418  \\
\hline 
mot, app, hiou, confidence  & 72.938	& \color{red}75.047 &	\color{red}58.695  \\
\hline
\textbf{Hadamard}&\multicolumn{3}{|c|}{ } \\
\hline
mot (iou) & 72.722 & 73.794 & 57.815 \\
\hline 
mot, app  & 72.747 & 74.155 & 58.091  \\
\hline 
mot, app, hiou  & 72.724 & 73.396 &	57.547  \\
\hline 
mot, app, hiou, confidence  & 72.7 &	73.271 & 57.432  \\
\hline
\end{tabular}
\label{MOT20}
\end{center}
\end{table}

\subsubsection{DanceTrack}
The comparison of different fusion methods on validation set of DanceTrack is given in Table~\ref{DanceTrack}. As can be seen in this table, the minimum fusion method has outstanding overall performance, where the highest values of HOTA and IDF1 are obtained. The highest value of MOTA is obtained using the hadamard fusion method. The overall performance of the KF gating fusion method is lower on the DanceTrack dataset. In general, the fusion of the appearance information with the motion information improves the tracking performance when using all the fusion methods. However, integrating the weak clues decreases the performance when using the minimum and hadamard fusion methods, similar to on MOT17 and MOT20 datasets. Incorporating height-IoU decreases the tracking performance when using the weighted sum based on IoU and KF gating fusion methods; however, incorporating confidence information improves the performance, as shown in Table~\ref{DanceTrack}.

\begin{table}[htbp]
\caption{Evaluation on DanceTrack validation dataset. Motion distance (mot) is computed using Mahalanobis distance for KF gating fusion method and using IoU for the other methods. The first and second highest values are highlighted by $\color{red}{\text{red}}$ and $\color{blue}{\text{blue}}$, respectively.}
\begin{center}
\begin{tabular}{|c|c|c|c|}
\hline
\textbf{ }&\multicolumn{3}{|c|}{\textbf{DanceTrack}} \\
\cline{2-4} 
\textbf{Minimum} &  \textbf{\textit{MOTA}}& \textbf{\textit{IDF1}}& \textbf{\textit{HOTA}} \\
\hline 
mot (iou) & 88.068 & 53.838	& 52.412  \\
\hline 
mot, app  & 88.168 & \color{red}60.105	& \color{red}58.474  \\
\hline 
mot, app, hiou  & 88.06 & \color{blue}57.04 & \color{blue}56.016   \\
\hline 
mot, app, hiou, confidence  & 87.922	& 53.538 & 54.47 \\
\hline
\textbf{Weighted-sum}&\multicolumn{3}{|c|}{ } \\   
\cline{2-4} 
\hline
mot (iou) & 88.068 &	53.838 & 52.412 \\
\hline 
mot, app  & 88.126 & 56.45 & 55.026   \\
\hline 
mot, app, hiou  & 88.118 & 54.393 & 54.066 \\
\hline 
mot, app, hiou, confidence  & \color{blue}88.172 & 55.664 & 55.70 \\ 
\hline
\textbf{KF-gating}&\multicolumn{3}{|c|}{ } \\
\cline{2-4} 
\hline
mot (mahalanobis) &  83.795 & 33.186 & 35.807  \\
\hline 
mot, app  &  86.419	& 47.098 &	49.08 \\
\hline 
mot, app, hiou  & 86.696 & 45.857 & 48.021  \\
\hline 
mot, app, hiou, confidence  & 86.716	& 46.393 & 48.695 \\
\hline
\textbf{Hadamard}&\multicolumn{3}{|c|}{ } \\
\hline
mot (iou) & 88.068 &	53.838 & 52.412 \\
\hline 
mot, app  & \color{red}88.228 & 56.039 & 54.938 \\
\hline 
mot, app, hiou  & 88.106 &	53.107 & 52.237  \\
\hline 
mot, app, hiou, confidence  & 88.038	& 52.996 & 52.07 \\
\hline
\end{tabular}
\label{DanceTrack}
\end{center}
\end{table}

%
%

\begin{table}[htbp]
\caption{Evaluation on MOT17 validation dataset for comparison of IoU and Mahalanobis distance for the second association of the KF gating fusion method. The first and second highest values are highlighted by $\color{red}{\text{red}}$ and $\color{blue}{\text{blue}}$, respectively.}
\begin{center}
\begin{tabular}{|c|c|c|c|}
\hline
\textbf{ }&\multicolumn{3}{|c|}{\textbf{MOT17}} \\
\cline{2-4} 
\textbf{IoU} &  \textbf{\textit{MOTA}}& \textbf{\textit{IDF1}}& \textbf{\textit{HOTA}} \\
\hline 
mot  &  76.749 & 69.256 &  60.581 \\
\hline 
mot, app  & 78.035 &  81.75 &  68.889 \\
\hline 
mot, app, hiou  & 78.072  &  \color{blue}82.249 &  \color{blue}69.29 \\
\hline 
mot, app, hiou, confidence  &  78.003 & \color{red}82.322 &  \color{red}69.366 \\
\hline
\textbf{Mahalanobis}&\multicolumn{3}{|c|}{ } \\
\hline
mot  &  76.708 &  68.453 & 60.158  \\
\hline 
mot, app  & \color{red}78.189 & 81.145 & 68.708 \\
\hline 
mot, app, hiou  & 78.115  & 81.613 & 69.02 \\
\hline 
mot, app, hiou, confidence  & \color{blue}78.128  & 81.567 & 69.101  \\
\hline
\end{tabular}
\label{IoUvsMahalanobis}
\end{center}
\end{table}

\section{Conclusion} \label{Conclusion}

In this paper, we investigate four commonly used different fusion methods for associating detections to tracklets in multi-object visual tracking by considering strong cues such as motion and appearance information as well as weak cues such as height intersection-over-union (height-IoU) and tracklet confidence information. The fusion methods include minimum, weighted sum based on IoU, Kalman filter (KF) gating, and hadamard product of costs due to the different cues. For computing confident cost, we elegantly incorporate tracklet confidence state into the state vector representation of the KF. Through extensive experiments on validation sets of MOT17, MOT20 and DanceTrack datasets, we find out that the different fusion methods have their own pros and cons. The minimum fusion method works reasonably well when fusing motion and appearance information; however, its performance decreases when incorporating weak cues. The incorporation of weak cues also degrades the tracking performance when using the hadamard fusion method. The tracking performance increases when using weak cues along with strong cues when using weighted sum and KF gating fusion methods though the performance of the KF gating fusion method is generally lower on DanceTrack dataset. Hence, the weighted sum based on IoU is more favourable when using weak cues along with the strong cues. We hope that this investigative work helps the computer vision research community to use the right fusion method with given cues for data association in multi-object visual tracking.

\bibliographystyle{IEEEtran}
\bibliography{refs}


\end{document}